\PassOptionsToPackage{prologue,dvipsnames}{xcolor}

\documentclass[sigconf]{acmart}
\newcommand{\proposed}{\textsf{AMOLE}}

\usepackage{graphicx}
\usepackage{subfiles}
\usepackage{enumitem}
\usepackage{multirow}
\usepackage{booktabs} 
\usepackage[dvipsnames]{xcolor}

\usepackage{amsmath}
\usepackage{mathtools}
\usepackage{amsthm}
\usepackage{pifont}
\usepackage{subfiles}

\newcommand{\cmark}{\textcolor{blue}{\ding{51}}}%
\newcommand{\xmark}{\textcolor{red}{\ding{55}}}%

\AtBeginDocument{%
  }

\copyrightyear{2024}
\acmYear{2024}
\setcopyright{rightsretained}
\acmConference[CIKM '24]{Proceedings of the 33rd ACM International Conference on Information and Knowledge Management}{October 21--25, 2024}{Boise, ID, USA}
\acmBooktitle{Proceedings of the 33rd ACM International Conference on Information and Knowledge Management (CIKM '24), October 21--25, 2024, Boise, ID, USA}
\acmDOI{10.1145/3627673.3679607}
\acmISBN{979-8-4007-0436-9/24/10}

\begin{document}

\title[Augmentation Strategies for Molecule Language Models]{Vision Language Model is NOT All You Need: \\ Augmentation Strategies for Molecule Language Models}



\author{Namkyeong Lee}
\affiliation{%
  \institution{KAIST}
  \city{Daejeon}
  \country{S. Korea}
}
\email{namkyeong96@kaist.ac.kr}
\authornote{Work done while the author was a visiting Ph.D. student at UIUC.}

\author{\mbox{Siddhartha Laghuvarapu}}
\affiliation{%
  \institution{UIUC}
  \city{Urbana, IL}
  \country{USA}
}
\email{sl160@illinois.edu}

\author{Chanyoung Park}
\affiliation{%
  \institution{KAIST}
  \city{Daejeon}
  \country{S. Korea}
}
\email{cy.park@kaist.ac.kr}
\authornote{~~Corresponding Author.}

\author{Jimeng Sun}
\authornotemark[2]
\affiliation{%
  \institution{UIUC}
  \city{Urbana, IL}
  \country{USA}
}
\email{jimeng@illinois.edu}

\renewcommand{\shortauthors}{Lee et al.}

\begin{abstract}
    Recently, there has been a growing interest among researchers in understanding molecules and their textual descriptions through molecule language models (MoLM). 
    However, despite some early promising developments, the advancement of MoLM still trails significantly behind that of vision language models (VLM).
    This is because unique challenges exist apart from VLM in the field of MoLM due to 1) a limited amount of molecule-text paired data and 2) missing expertise that occurred due to the specialized areas of focus among the experts.
    To this end, we propose \proposed, which 1) augments molecule-text pairs with structural similarity preserving loss, and 2) transfers the expertise between the molecules.
    Specifically, \proposed~enriches molecule-text pairs by sharing descriptions among structurally similar molecules with a novel structural similarity preserving loss. 
    Moreover, we propose an expertise reconstruction loss to transfer knowledge from molecules that have extensive expertise to those with less expertise.
    Extensive experiments on various downstream tasks demonstrate the superiority of \proposed~in comprehending molecules and their descriptions, highlighting its potential for application in real-world drug discovery.
    The source code for \proposed~is available at ~\textcolor{magenta}{\url{https://github.com/Namkyeong/AMOLE}}.
\end{abstract}


\maketitle

\section{Introduction}
Machine learning (ML) techniques have recently brought significant advancements to molecular science \cite{zhang2023artificial,wang2023scientific,lee2023conditional}, a field that has traditionally relied on theory, experimentation, and computationally intensive simulations. 
Particularly, Language Models (LMs), inspired by their recent successes in various fields, have rapidly become popular in molecular science, driven by the wealth of literature available in this domain.
Notably, the conventional method of representing molecules as strings, such as SMILES strings \citep{weininger1988smiles}, facilitates the integration of two different modalities, i.e., text and molecules, into a single LM.
Then, following the masked language modeling \citep{devlin2018bert}, \citet{zeng2022deep} train the model on masked SMILES and text from the scientific literature.
Moreover, inspired by T5 models \citep{raffel2020exploring}, training the Molecule Language Model (MoLM) with multiple tasks and fine-tuning the downstream task has been proposed \citep{edwards2022translation,pei2023biot5,christofidellis2023unifying}.
However, all these models require molecules to be represented in a string format, like SMILES, to be understood by language models.

As an alternative approach, one can treat molecules and language as multiple modalities, following the recent success of the vision language model (VLM).
For instance, several recent works \citep{edwards2021text2mol,liu2023multi,seidl2023enhancing} have proposed using separate encoders for each modality and training the paired modalities to be similar in the representation space using contrastive learning, especially inspired by the CLIP model \citep{radford2021learning}. 
These models successfully capture complementary information between different modalities by learning a joint representation space that can be utilized for various downstream tasks such as cross-modal retrieval and molecular property prediction.

\begin{figure}[t]
    \centering
    \includegraphics[width=0.99\columnwidth]{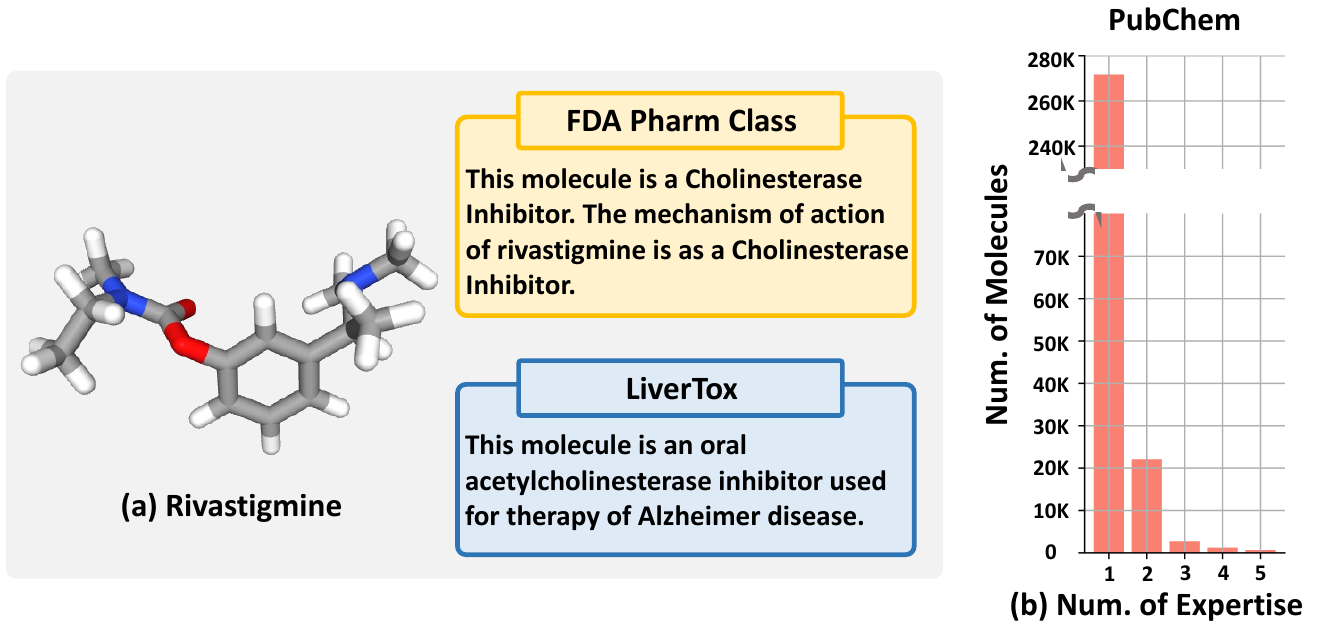} 
    \vspace{-2ex}
    \caption{(a) Rivastigmine's textual descriptions from various experts. (b) The majority of molecules in the PubChem database have only one description provided by an expert.}
    \label{fig: intro}
\end{figure}

Despite promising early strides, the progress of MoLM lags far behind its VLM counterparts due to the scarcity of molecule-text paired data, both in \textit{quantity and expertise}.
First, in terms of quantity, the VLM community largely follows the viewpoint that scale is everything, as image-text pairs are widely available on the web \citep{wang2023too}.
Pre-training VLM on the massive amounts of crawled image-text pairs, ranging from tens of thousands to billions, consistently leads to significant performance gains in various downstream tasks \citep{jia2021scaling,yu2022coca}. 
In contrast, MoLM faces a bottleneck due to the \textit{limited amount of molecule-text paired data} available, 
since costly domain knowledge and significant time investment in wet lab experiments are required for generating molecule-text pairs.
Therefore, although it is typical to expand image-text pairs by crawling the web, augmenting molecule-text pairs is not straightforward.

Second, in terms of expertise, each molecule has various descriptions provided by different experts, each focusing on its unique areas of expertise.
For example, in the case of a molecule named Rivastigmine (Figure \ref{fig: intro} (a)), FDA explains its function as a cholinesterase inhibitor, whereas LiverTox \citep{hoofnagle2013livertox} focuses on its effectiveness on Alzheimer's disease. 
However, due to the specialized areas of the experts, they typically restrict their knowledge to a selective group of molecules, resulting in numerous molecules having \textit{missing expertise} across various experts.
Specifically, our examination of the largest molecule database, PubChem \cite{kim2021pubchem}, as illustrated in Figure \ref{fig: intro} (b), reveals that only 27K out of 299K molecules have descriptions from multiple experts. 
Consequently, the remaining 272K molecules are each documented with a single expert description, which may lack comprehensive expertise about the molecule.


In this paper, we propose \proposed~that addresses the unique challenges faced in MoLM by \textsf{A}ugmenting \textsf{MOL}ecule-text pair and transferring \textsf{E}xpertise between the molecules.
To overcome the lack of molecule-text paired data, we utilize the idea that molecules with similar structures have similar properties, which is grounded in the well-established biochemical principle \citep{martin2002structurally}. 
Specifically, we propose to augment molecule-text pairs by sharing descriptions among structurally similar molecules.
However, naively treating augmented molecule-text pairs as positive pairs may not be beneficial, as the description is not specifically tailored for structurally similar molecules.
To address this issue, instead of directly guiding the model to align augmented pairs closely to the representation, 
we introduce a novel loss function which instructs the model to align the augmented pairs more accurately based on the structural similarity of the molecules sharing the description.


Furthermore, to address the missing expertise issue, we utilize the fact that different areas of expertise are interrelated, allowing us to infer additional expertise based on one known area.
As an example, in Figure \ref{fig: intro} (a), since the Cholinesterase inhibitor is widely known to improve communication between nerve cells by increasing levels of Acetylcholine in the nervous system \citep{grossberg2003cholinesterase}, one could infer Livertox's expertise about Alzheimer's disease from FDA descriptions.
To this end, we propose to transfer the expertise acquired from molecules with extensive descriptions to those with less description.
Specifically, given a molecule that possesses descriptions from multiple experts, we train the model to reconstruct one description from another, thereby enhancing the ability to deduce expertise from one expert using information from another. 
With our proposed training strategy, the model behaves as if it has access to abundant expertise, even if the model faces the molecule with missing expertise.

In this paper, we make the following contributions:
\begin{itemize} [leftmargin=.1in]
    \item To increase the \textit{limited amount} of molecule-text paired data, we propose to selectively share descriptions among molecules with a novel loss function based on their structural similarity.
    \item To address the issue of \textit{missing expertise}, we propose to transfer the expertise between molecules by enhancing the model's ability to reconstruct one description from another.
    \item Extensive experiments including two novel and practical tasks; zero-shot question and answering, and zero-shot virtual screening, demonstrate the superiority and potential applicability of \proposed~in real-world drug discovery process.
\end{itemize}
\vspace{-0.5ex}
\noindent To the best of our knowledge, this is the first paper addressing the data scarcity in MoLM in terms of quantity and expertise.

\section{Related Works}
\noindent\textbf{MoLM with a Single Language Model.}
Thanks to the wealth of literature and the traditional string-based representation of molecules, such as SMILES, LMs have been applied to the domain of molecular science.
Drawing inspiration from the masked language model approach used in BERT training \citep{devlin2018bert}, KV-PLM \citep{zeng2022deep} proposes to train LMs by reconstructing masked SMILES and textual data.
Moreover, MolT5 \citep{edwards2022translation} proposes to pre-train the LMs with the ``replace corrupted spans'' objective \citep{raffel2020exploring} on both SMILES string and textual data, followed by fine-tuning for tasks such as molecule captioning and generation.
\citet{pei2023biot5} and \citet{christofidellis2023unifying} extend MolT5 with various pre-training tasks, such as protein FASTA reconstruction and chemical reaction prediction.
However, these models rely on string-based representations of molecules, e.g., SMILES, which are broadly recognized for their lack of topology awareness \citep{rong2020self}.
Furthermore, merging two modalities into a single model prevents the adoption of existing superior pre-trained models tailored for each modality \citep{liu2023multi}.

\smallskip
\noindent\textbf{MoLM with Multi-Modal Contrastive Learning.}
Inspired by the recent success of VLM, researchers have started to conceptualize molecules and text as separate modalities, particularly by adopting separate encoders for each modality.
As a pioneering work, Text2Mol \citep{edwards2021text2mol} proposes training separate encoders for molecular graph and textual description with cross-modal contrastive learning.
Following this, CLAMP \citep{seidl2023enhancing} suggests employing contrastive learning for predicting activities based on the textual description of the task. 
Furthermore, MoleculeSTM \citep{liu2023multi} develops the largest multi-modal dataset sourced from the PubChem database for cross-modal contrastive learning applications. 
It is essential to recognize that the distinction between these models is based on their architectural design and the data used for training rather than the training loss itself. 
On the other hand, MoMu \cite{su2022molecular} introduces an intermolecular contrastive loss along with random molecular augmentations like node dropping, which may lead to chemically invalid structures \cite{lee2022augmentation}.
Unlike previous works, \proposed~concentrates on overcoming the specific hurdles uniquely encountered in MoLM: the \textit{lack of abundance and expertise in molecule-text pairs}, through innovative training loss strategies.

\smallskip
\noindent\textbf{MoLM with Other Multi-Modal Learning.}
While the majority of MoLM relies on CLIP style architecture \cite{radford2021learning}, there have been other approaches for integrating molecule text through a novel model architecture.
As an example, GIMLET \citep{zhao2023gimlet} suggests encoding graph structure and instructional text directly, without separate graph encoding modules, by utilizing generalized position embeddings. 
Additionally, drawing inspiration from BLIP-2 \citep{li2023blip} in VLM, MolCA \citep{liu2023molca} implements a three-stage training pipeline. 
The initial stage focuses on training the Q-Former to extract molecule representations relevant to the text, and the second stage targets aligning molecular graphs with texts through language modeling. 
The final stage involves fine-tuning the model for downstream generation tasks.
Note that these works are not directly related to our research, as we focus on introducing novel training loss designed for molecule-text pair augmentation and expertise transfer focusing on CLIP-style architecture.

\section{Preliminaries}
\subsection{Problem Statement}
\textbf{Notations.} 
Let $\mathcal{G} = (\mathcal{V}, \mathcal{E})$ represent a molecular graph with atoms $\mathcal{V}$ as nodes and the edges $\mathcal{E}$ given by covalent bonds.
Moreover, we have a set of textual descriptions $\mathcal{T} = \{t^1, \ldots, t^{N}\}$ regarding the molecule $\mathcal{G}$ from $N$ different experts, such as FDA and LiverTox, in Figure \ref{fig: intro} (a), each of which details various attributes of the molecule.
Note that the number of descriptions for each molecule varies, i.e., $N_{i}$ depends on molecule $\mathcal{G}_{i}$.

\smallskip
\noindent \textbf{Task Description.} 
Given a molecular graph $\mathcal{G}$ with its textual description $t \in \mathcal{T}$, we aim to train encoders $f_{\text{mol}}$ and $f_{\text{text}}$ that produce a molecule representation $z_{\mathcal{G}} \in \mathbb{R}^{D}$ and a textual representation $z_{t} \in \mathbb{R}^{D}$, respectively.
More specifically, we aim to obtain the encoders that produce a generalized representation of molecules and text, that can be utilized for a wide range of downstream tasks, such as zero-shot cross-modal retrieval and zero-shot virtual screening.

\subsection{Tanimoto Similarity}
One traditional way of representing a molecule is through fingerprints, a series of binary bits indicating the presence or absence of specific substructures \cite{rogers2010extended}.
The Tanimoto similarity is a widely accepted criterion for calculating the similarity between two molecules \cite{bajusz2015tanimoto} based on the fingerprints.
Specifically, for the molecules $\mathcal{G}_i$ and $\mathcal{G}_j$ that are represented with the fingerprints $\mathtt{fp}_i$ and $\mathtt{fp}_j$, respectively, the Tanimoto similarity is calculated as follows:
\begin{equation}
s_{ij} = \frac{|\mathtt{fp}_i \cap \mathtt{fp}_j|}{|\mathtt{fp}_i| + |\mathtt{fp}_j| -  |\mathtt{fp}_i \cap \mathtt{fp}_j|}.
\label{eq: Tanimoto}
\end{equation}
Intuitively, the Tanimoto similarity takes both common and distinct substructures between two molecules into account, thereby offering an assessment of their structural similarity.

\subsection{Molecule-Text Contrastive Learning}
Previous works \citep{liu2023multi,su2022molecular} have introduced multi-modal contrastive learning to obtain encoders that establish qualified joint space between a molecule $\mathcal{G}$ and its corresponding text $t$.
This approach ensures that paired molecules and texts are closely aligned in the representation space, while unpaired ones remain distant.
Specifically, given a molecule $\mathcal{G}$ and its corresponding text $t$, we first obtain the molecule and text representations as follows:
$z_{\mathcal{G}} = f_{\text{mol}}(\mathcal{G})$ and $z_{t} = f_{\text{text}}(t)$.
Then, the model is trained with the following Noise-Contrastive Estimation (InfoNCE) \citep{oord2018representation} loss:
\begin{equation}
\begin{split}
    \mathcal{L}_{\text{InfoNCE}} = - \frac{1}{2} \Bigl\{ \log\frac{\exp(\mathtt{sim}(z_{\mathcal{G}}, z_{t})/\tau)}{\sum_{t'}{\exp(\mathtt{sim}(z_{\mathcal{G}}, z_{t'})/\tau)}} \\ 
    + \log\frac{\exp(\mathtt{sim}(z_{t}, z_{\mathcal{G}})/\tau)}{\sum_{\mathcal{G}'}{\exp(\mathtt{sim}(z_{t}, z_{\mathcal{G}'})/\tau)}} \Bigr\},
\end{split}
\label{eq: InfoNCE}
\end{equation}
where $t'$ and $\mathcal{G}'$ represent all the texts and molecules within the batch, respectively, $\mathtt{sim}(\cdot, \cdot)$ indicates the cosine similarity between the representations, and $\tau$ denotes the temperature hyperparameter, which controls the strength of penalties on negative samples \cite{wang2021understanding}.

\section{Methodology}
\begin{figure}[t]
    \centering
    \includegraphics[width=1.02\columnwidth]{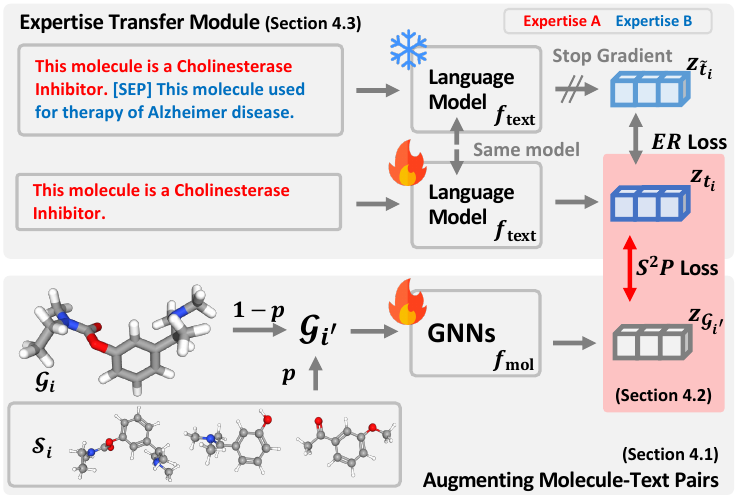} 
    \caption{Overall model architecture of~\proposed.}
    \label{fig: model}
\end{figure}

In this section, we introduce our proposed method, called \proposed, a novel molecule-text contrastive learning approach that addresses the two unique challenges faced in MoLM, i.e., scarcity of molecule-text paired data in terms of quantity and expertise. 
We first explain how to increase the number of molecule-text pairs (\textbf{Section \ref{sec: Augmenting Molecule-Text pairs}}). 
Then, we introduce our novel loss function that aligns augmented pairs more precisely in the representation space (\textbf{Section \ref{sec: Structural Similarity Preserving Loss}}).
Finally, we show how to transfer the expertise between molecules (\textbf{Section \ref{sec: Expertise Transfer Module}}).
Figure \ref{fig: model} presents the overall model architecture of \proposed.

\subsection{Augmenting Molecule-Text pairs}
\label{sec: Augmenting Molecule-Text pairs}
In the VLM community, it is well-recognized that a substantial quantity of image-text pairs is crucial for training the model \citep{jia2021scaling,yu2022coca}, which can be easily accomplished by crawling the image-text pairs from the Web.
However, molecule-text pairs are typically produced by specialized experts, making it more challenging to augment the data with external sources such as Web.
To this end, we propose to augment the molecule-text pair by sharing the textual description among molecules within the existing data.
The key idea here is to share textual descriptions among molecules with structural resemblances, based on the well-established biochemical principle that structurally similar molecules often display analogous biological activities \citep{martin2002structurally}.
In other words, we posit that molecules with similar structures can share textual descriptions that detail their biochemical properties.

Specifically, we begin by computing all the pairwise structural similarity, i.e., Tanimoto similarity (Eq. \ref{eq: Tanimoto}), between the molecules in the training set.
Then, given the similarity information of a molecule $\mathcal{G}_{i}$ to other molecules, 
we identify the top $k$ molecules that exhibit the highest similarity to $\mathcal{G}_{i}$. We represent these molecules as the set $\mathcal{S}_{i}$.
During training, a molecule $\mathcal{G}_{i'}$ is randomly selected from the set $\mathcal{S}_{i}$ and substituted for the original molecule $\mathcal{G}_{i}$ according to a predetermined probability $p$; otherwise, it is kept identical to the original molecule $\mathcal{G}_{i}$.
By doing so, the textual description $t_{i}$ of molecule $\mathcal{G}_{i}$ is shared among structurally similar $k$ molecules within the set $\mathcal{S}_{i}$, generating a new molecule-text pair ($\mathcal{G}_{i'}$, $t_i$). 
As a result, this approach effectively expands the initial $n$ molecule-text pairs to $n \cdot k$ pairs without costly wet-lab experiments or the expense of labeling with domain expertise.
Note that $\mathcal{S}_i$ for each molecule $\mathcal{G}_i$ can be readily defined before the model training, thus do not require any additional computational cost during the training process.

\subsection{Structural Similarity Preserving Loss}
\label{sec: Structural Similarity Preserving Loss}
While we increase the number of pairs by sharing the descriptions, training the model with augmented pairs is challenging since the shared description $t_i$ is not specifically written for the substituted molecule $\mathcal{G}_{i'}$.
Consequently, employing traditional InfoNCE loss in Eq. \ref{eq: InfoNCE}, which strictly encourages the positive pairs to be close in the representation space, might not be suitable for training as the augmented pairs may not truly be positive.

To address this issue, we propose structural similarity preserving ($S^{2}P$) loss, designed to preserve molecules' structural similarity in molecule-text joint space.
That is, given a text $t_{i}$ and a molecule $\mathcal{G}_{i'}$, instead of directly defining the pair as positive, we propose to utilize the Tanimoto similarity $s_{ii'}$ between $\mathcal{G}_{i}$ and $\mathcal{G}_{i'}$ as a pseudo label for contrastive learning.
In particular, for a given molecule $\mathcal{G}_{i}$, we compile a set of $s_{ij'}$ values, where $j'=1, \ldots, N_{\text{batch}}$ represents the range of molecules within a batch.
Then, the pseudo label is calculated as follows:
\begin{equation}
    y_{ij'}^{t \rightarrow m} = \frac{\exp({s_{ij'}}/\tau_{1})}{\sum_{k'=1}^{N_{\text{batch}}}\exp({s_{ik'}}/\tau_{1})},
\label{eq: mol2text pseudo label}
\end{equation}
where $\tau_{1}$ is a temperature hyperparameter.
Then, we make a prediction on the pseudo label based on the similarity between text $t_i$ and molecule $\mathcal{G}_{j'}$ in representation space as follows:
\begin{equation}
    \hat{y}_{ij'}^{t \rightarrow m} = \frac{\exp(\texttt{sim}(z_{t_i}, z_{\mathcal{G}_{j'}})/\tau_{2})}{\sum_{k'=1}^{N_{\text{batch}}}\exp(\texttt{sim}(z_{t_i}, z_{{\mathcal{G}}_{k'}})/\tau_{2})},
\label{eq: mol2text pseudo label}
\end{equation}
where $\tau_{2}$ is a temperature hyperparameter.
We generate pseudo labels by applying softmax normalization to structural similarity since the loss aims to approximate the representation similarity $\texttt{sim}(z_{t_i}, z_{\mathcal{G}_{j'}})$ between a molecule and text to structural similarity $s_{ij'}$.
Then \proposed~is trained with the following cross-entropy loss:
\begin{equation}
    \mathcal{L}^{t \rightarrow m}_{S^{2}P} = -\frac{1}{N_{Batch}}\sum_{i=1}^{N_{batch}}\sum_{j'=1}^{N_{batch}}{y_{ij'}^{t \rightarrow m}\log\hat{y}_{ij'}^{t \rightarrow m}}.
\label{eq: mol2text cross-entropy}
\end{equation}
By doing so, the structural similarity between molecules $\mathcal{G}_i$ and $\mathcal{G}_{i'}$ instructs the model on how to align the text $t_i$ and molecule $\mathcal{G}_{i'}$ closely in the representation space, enabling the selective sharing of descriptions based on structural similarities.
For the symmetricity, we compute $\mathcal{L}^{m \rightarrow t}_{S^{2}P}$ and derive final structural similarity preserving loss: $\mathcal{L}_{S^{2}P} = \mathcal{L}^{t \rightarrow m}_{S^{2}P} + \mathcal{L}^{m \rightarrow t}_{S^{2}P}$.


\subsection{Expertise Transfer Module}
\label{sec: Expertise Transfer Module}
On the other hand, experts in molecular science often have a specific area of focus, resulting in some molecules lacking comprehensive expertise from different specialists.
To address this issue of missing expertise, we suggest transferring the expertise gained from molecules with extensive descriptions to those with less description. 
Specifically, we use a molecule $\mathcal{G}_{i}$ with a set of descriptions $\mathcal{T}_{i}$ obtained from various experts. We train the language model $f_{\text{text}}$ to reconstruct the description $t_{i'} \in \mathcal{T}_{i}$ from a given description $t_{i}$.
This approach allows the language model $f_{\text{text}}$ to become skilled at inferring expertise from one institution based on another. 
As a result, it behaves as if it has access to abundant expertise even when dealing with molecules with missing expertise.

However, reconstructing the description $t_{i'}$ in text space poses a challenge, as our model is designed to learn a qualified representation space without a decoder structure.
To this end, we propose a novel expertise reconstruction ($ER$) loss, which guides the model to reconstruct the description within the representation space rather than directly in the text space.
More formally, when we have a textual description $t_{i}$ and another associated description $t_{i'}$ for a molecule $\mathcal{G}_{i}$, we formulate the reconstruction target $\Tilde{t}_{i}$ by concatenating the two descriptions with \textsf{[SEP]} token, i.e.,  $\Tilde{t}_{i} = $ $t_{i}$ \textsf{[SEP]} $t_{i'}$. 
Then, the model is trained to minimize the L2 distance between the textual description $t_{i}$ and reconstructed target $\Tilde{t}_{i}$ as follows:
\begin{equation}
    \mathcal{L}_{ER} = -\frac{1}{N_{Batch}}\sum_{i=1}^{N_{batch}}{\|f_{\text{text}}(t_{i}) - \text{SG}(f_{\text{text}}(\Tilde{t}_{i})) \|^{2}_{2}},
\label{eq: ER Loss}
\end{equation}
where SG denotes the stop-gradient operation, which halts the propagation of gradients when inputs are structured as $\Tilde{t_{i}}$, thereby preventing the model from acquiring degenerate solutions by disregarding the text following the \textsf{[SEP]} token.
It is also worth noting that since $f_{\text{text}}$ is a language model, it can accommodate texts of varying lengths.
Moreover, as illustrated in Figure \ref{fig: intro} (b), molecules described by more than two texts are rare, and this scarcity could reduce the effectiveness of the $ER$ Loss.
To mitigate this problem, we sample the $N_{batch}$ number of molecules that have more than two descriptions and apply the $ER$ loss to those molecules.
In conclusion, with the training loss, the model behaves as if there exists extensive expertise available, enabling reliable predictions even when a detailed description of the molecule is lacking.

\subsection{Model Training}
Finally, \proposed~is trained by jointly optimizing two losses, i.e., structural similarity preserving loss and expertise reconstruction loss, as follows:
\begin{equation}
    \mathcal{L} = \mathcal{L}_{S^{2}P} + \alpha \cdot \mathcal{L}_{ER},
\label{eq: Final Loss}
\end{equation}
where $\alpha$ denotes the hyperparameter for controlling the weight of the expertise reconstruction loss.

\section{Experiments}
\subsection{Experimental Setup}

\textbf{Pretraining Dataset.}
We pre-train \proposed~with PubChem database \cite{kim2021pubchem}, which is one of the most extensive public molecular databases available.
PubChem database consists of multiple data sources including DrugBank \cite{wishart2018drugbank}, CTD \cite{davis2021comparative}, PharmGKB \cite{thorn2013pharmgkb}, and more.
\footnote{Please refer to the following URL for more details on the data sources included in PubChem database: \url{https://pubchem.ncbi.nlm.nih.gov/sources/}.}
Our pre-training dataset is compiled using the preprocessing script provided in the repository of the previous work \cite{liu2023multi}. 
However, due to regular updates to the PubChem database, our dataset varies slightly from the previous work, comprising a total of 299K unique molecules and 336K molecule-text pairs.
During preprocessing, we consolidate each expertise into a unified description, meaning that each description of an individual molecule originates from distinct expertise.
As shown in Figure \ref{fig: intro} (b), it is evident that the majority of molecules have a singular description, indicating a lack of comprehensive expertise across various experts for numerous molecules.
On average, molecules are associated with 1.115 descriptions, with a maximum of 17 descriptions and a minimum of one.
Moreover, each description in training data consists of 17.62 words on average, with a maximum of 874 words and a minimum of one.

\smallskip
\noindent \textbf{Downstream Tasks.}
After training the model using the PubChem database, we assess its performance on \textbf{four} distinct downstream tasks: 1) Zero-Shot Cross-Modal Retrieval, 2) Zero-Shot Question and Answering, 3) Molecular Property Prediction, and 4) Zero-Shot Virtual Screening.
While tasks such as 1) Zero-Shot Cross-Modal Retrieval and 3) Molecular Property Prediction are commonly explored in previous studies \cite{liu2023multi},
we propose novel tasks in this paper: 2) Zero-Shot Question and Answering, and 4) Zero-Shot Virtual Screening tasks that are essential for evaluating the model's performance on fine-grained knowledge extraction and potential for application in real-world drug discovery, respectively.

\smallskip
\noindent \textbf{Implementation Details.}
Following previous work \cite{liu2023multi}, we use graph isomorphism network (GIN) \cite{xu2018powerful} architecture as a molecular encoder $f_{\text{mol}}$, which has been widely used as the backbone model in recent graph self-supervised learning works \cite{hu2019strategies,liu2021pre}.
Moreover, we use BERT architecture \cite{devlin2018bert} for the text encoder $f_{\text{text}}$.
One key benefit of treating molecules and text as distinct modalities is the opportunity to leverage powerful pre-trained models specifically designed for each modality. 
Following a previous work \cite{liu2023multi}, we employ a GraphMVP \cite{liu2021pre} pre-trained checkpoint for GIN model \cite{xu2018powerful} for our molecule encoder $f_{\text{mol}}$.
Moreover, for our text encoder $f_{\text{text}}$, we utilize a pre-trained SciBERT \cite{beltagy2019scibert}, which has been trained on a vast corpus of textual data from the biochemistry domains.

\smallskip
\noindent \textbf{Training Details.}
Our method is implemented on Python 3.7.16, PyTorch 1.10.1, and Torch-geometric 2.0.3. All experiments are conducted using an 80GB NVIDIA A100 GPU.
It takes 90 minutes per epoch for training, a total of 2700 minutes for training.

\smallskip
\noindent \textbf{Baseline Methods.}
We evaluate \proposed~against three single encoder models: MolT5 \cite{edwards2022translation}, BioT5 \cite{pei2023biot5}, and KV-PLM \cite{zeng2022deep}, all of which represent molecules as 1D SMILES strings.
While T5-based models are not designed for learning representations of molecules and text, we assess their capabilities by examining the hidden representations produced by the encoder model following a previous work \cite{seidl2023enhancing}.
Additionally, we compare \proposed~with two separate encoder models: MoMu \cite{su2022molecular} and MoleculeSTM \cite{liu2023multi}.
While MoMu depicts molecules as 2D graph structures, MoleculeSTM introduces two models that utilize both SMILES and 2D graph representations for molecules.
For the single encoder models, we utilize the checkpoints provided by the authors of the original papers.
However, for models with separate encoders, \textit{we independently pre-train models with the same pre-training data and model architecture for a fair comparison with the proposed}~\proposed.


\subsection{Zero-Shot Cross-Modal Retrieval}
\label{sec: Zero-Shot Cross-Modal Retrieval}

\begin{table}[t]
\centering
\caption{Model accuracy (\%) in zero-shot cross-modal retrieval task. The value within the brackets indicates the variance observed across five trials. Bold denotes the best performance.}
    \resizebox{0.99\linewidth}{!}{
    \begin{tabular}{lcccccccccc}
    \toprule
    & \multirow{2}{*}{\textbf{SMILES}} & \multirow{2}{*}{\textbf{Graph}} & & \multicolumn{3}{c}{\textbf{Given Molecule @ 20}} & & \multicolumn{3}{c}{\textbf{Given Text @ 20}} \\
    \cmidrule{5-7} \cmidrule{9-11}
    & & & & Descr. & Pharma. & ATC & & Descr. & Pharma. & ATC \\
    \midrule
    \multicolumn{8}{l}{\textbf{Single Encoder}} \\
    \midrule
    \multirow{2}{*}{MolT5} & \multirow{2}{*}{\cmark} & \multirow{2}{*}{\xmark} & & 5.06 & 6.80 & 6.48 & & 6.66 & 6.02 & 6.10  \\
    &  & & & \footnotesize{(0.44)} & \footnotesize{(0.28)} & \footnotesize{(0.25)} & & \footnotesize{(2.02)} & \footnotesize{(0.57)} & \footnotesize{(0.09)} \\
    \multirow{2}{*}{BioT5} & \multirow{2}{*}{\cmark} & \multirow{2}{*}{\xmark} & & 6.47 & 7.42 & 7.71 & & 6.02 & 7.36 & 6.78 \\
    & & & & \footnotesize{(0.13)} & \footnotesize{(0.52)} & \footnotesize{(0.16)} & & \footnotesize{(0.42)} & \footnotesize{(0.13)} & \footnotesize{(0.45)} \\
    \multirow{2}{*}{KV-PLM} & \multirow{2}{*}{\cmark} & \multirow{2}{*}{\xmark} & & 42.28 & 36.84 & 30.21 & & 45.64 & 37.93 & 33.22 \\
    & & & & \footnotesize{(3.29)} & \footnotesize{(0.53)} & \footnotesize{(0.40)} & & \footnotesize{(2.51)} & \footnotesize{(0.62)} & \footnotesize{(0.40)} \\
    \midrule
    \multicolumn{8}{l}{\textbf{Separate Encoder}} \\
    \midrule
    \multirow{2}{*}{MoMu} & \multirow{2}{*}{\xmark} & \multirow{2}{*}{\cmark} & & \textbf{97.39} & 77.82 & 51.34 & & 96.84 & 77.05 & 47.68 \\
    & & & & \footnotesize{(0.19)} & \footnotesize{(0.54)} & \footnotesize{(0.37)} & & \footnotesize{(0.17)} & \footnotesize{(0.28)} & \footnotesize{(0.34)} \\
    \multirow{2}{*}{MoleculeSTM} & \multirow{2}{*}{\cmark} & \multirow{2}{*}{\xmark} & & 96.70 & 77.28 & 52.36 & & 96.22 & 75.01 & 50.01 \\
    & & & & \footnotesize{(0.35)} & \footnotesize{(0.94)} & \footnotesize{(0.29)} & & \footnotesize{(0.29)} & \footnotesize{(0.49)} & \footnotesize{(0.40)} \\
    \multirow{2}{*}{MoleculeSTM} & \multirow{2}{*}{\xmark} & \multirow{2}{*}{\cmark} & & 95.87 & 79.21 & 52.70 & & 95.82 & 77.15 & 48.54 \\
     & & & & \footnotesize{(1.87)} & \footnotesize{(0.75)} & \footnotesize{(0.75)} & & \footnotesize{(0.37)} & \footnotesize{(0.74)} & \footnotesize{(0.49)} \\
    \midrule
    \multirow{2}{*}{\proposed} & \multirow{2}{*}{\xmark} & \multirow{2}{*}{\cmark} & & 96.48 & \textbf{81.46} & \textbf{54.76} & & \textbf{97.20} & \textbf{80.11} & \textbf{51.47} \\
     & & & & \footnotesize{(2.94)} & \footnotesize{(0.60)} & \footnotesize{(0.57)} & & \footnotesize{(0.26)} & \footnotesize{(0.42)} & \footnotesize{(0.56)} \\
    \bottomrule
    \end{tabular}}
    \label{tab: cross retrieval}
\end{table}

\noindent \textbf{Task Description.}
This task requires choosing an appropriate description from several alternatives for a specific molecule (\textbf{Given Molecule}) or retrieving the molecule that aligns with a particular description (\textbf{Given Text}). 
Following a previous work \cite{liu2023multi}, our experiments are carried out with 20 options, where one of the options is the matching counterpart, while the others are randomly selected from the dataset.
Then, the model performance is determined by its capacity to pinpoint the correct counterpart from the options provided, such as correctly matching the description to the given molecule or vice versa.
For the evaluation, we conducted five separate experiments, each featuring a distinct random selection of options, and we present both the mean and standard deviation of these experiments.
Moreover, \textit{we ensure that molecules appearing in the training dataset with identical canonical SMILES are excluded to avoid data leakage.}
We provide further details on the datasets used for evaluation in Appendix \ref{app: dataset zero-shot cross-modal retrieval}.

\smallskip
\noindent \textbf{Empirical Results.}
In Table \ref{tab: cross retrieval}, we have the following observations:
\textbf{1)} As T5-based models (i.e., MolT5 and BioT5) are not intended to build joint representation space between molecule and text, their performance was the worst among the models.
This suggests that T5-based models struggle to learn suitable representations and necessitate a costly fine-tuning process after the pre-training step.
\textbf{2)} Additionally, despite the SMILES-based MoleculeSTM having a significantly higher number of parameters compared to the graph-based model,
\footnote{For MoleculeSTM SMILES, we use ChemBERTa, which contains 83,450,880 parameters, and for MoleculeSTM Graph, we use GIN, which contains 1,885,206 parameters.}
SMILES and graph structure representations of molecules yield comparable results. 
This underscores the efficacy of graph-based representations in capturing molecular properties.
\textbf{3)} Within the graph-based models, it is evident that MoMu generally exhibits the lowest performance across all but one dataset.
This is largely due to the random augmentation of molecules, such as by dropping nodes, without taking the chemical validity of the molecules into account \cite{lee2022self}, which makes the model perform even worse than MoleculeSTM.
On the other hand, in the upcoming Ablation studies, we will demonstrate that augmenting molecule-text pairs with chemically valid molecules can consistently improve MoleculeSTM performance.
\textbf{4)} Overall, we find that \proposed~achieves the best results in five out of six datasets, illustrating its capability to understand and integrate two distinct modalities. 
Notably, the performance gap widens with increasingly challenging tasks (i.e., Pharma. and ATC, where overall model performance is lower), highlighting \proposed's intricate grasp of the modalities.


\begin{table}
\centering
\caption{Ablation studies results. }
    \vspace{-2ex}
    \resizebox{0.9\linewidth}{!}{
    \begin{tabular}{lccccccc}
    \toprule
    & Aug- & $S^2P$ & $ER$ & & \multirow{2}{*}{Descr.} & \multirow{2}{*}{Pharma.} & \multirow{2}{*}{ATC} \\
    & ment & Loss & Loss & &  &  &  \\
    \midrule
    MoleculeSTM & \xmark & \xmark & \xmark & & 95.85 & 78.18 & 50.62 \\
    Ablation 1 & \cmark & \xmark & \xmark & & 96.48 & 78.65 & 51.46 \\
    Ablation 2 & \cmark & \cmark & \xmark & & 96.65 & 80.47 & 51.55 \\
    \midrule
    \proposed & \cmark & \cmark & \cmark & & \textbf{96.84} & \textbf{80.79} & \textbf{53.12} \\
    \bottomrule
    \end{tabular}}    
    \label{tab: ablation studies}
    \vspace{-2ex}
\end{table}

\begin{figure}[t]
    \centering
    \includegraphics[width=0.9\columnwidth]{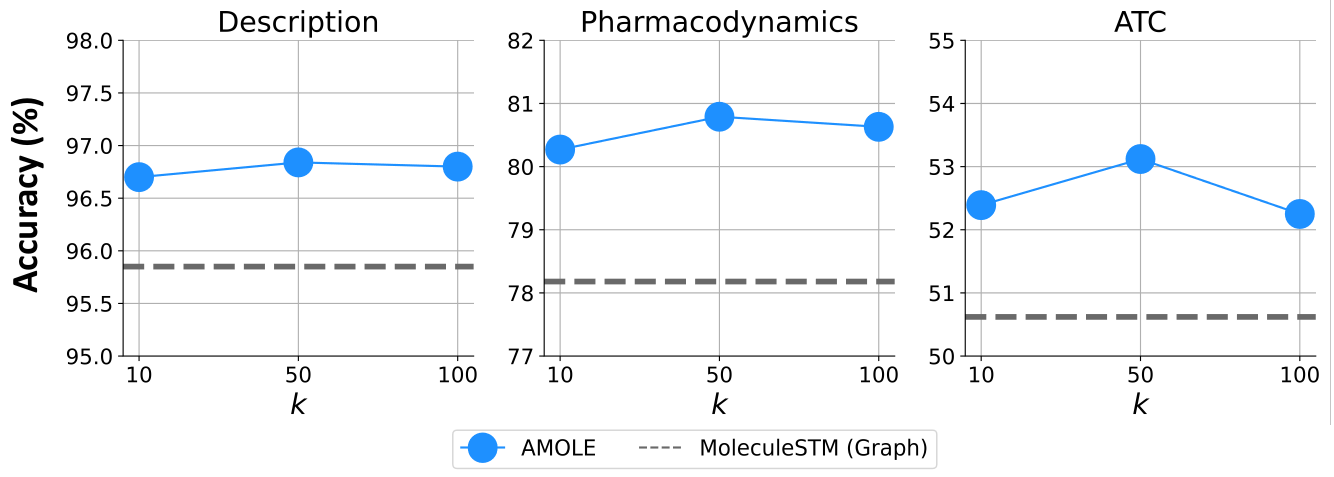} 
    \vspace{-2ex}
    \caption{Sensitivity analysis on $k$.}
    \label{fig: sensitivity analysis k}
    \vspace{-3ex}
\end{figure}

\smallskip
\noindent \textbf{Ablation Studies.}
Now, we empirically evaluate the impact of each component in \proposed~by sequentially removing them one at a time.
We assess the effectiveness of each component by calculating the average performance across cross-retrieval tasks, including retrieving among 20 texts given a molecule and retrieving among 20 molecules given a text.
In Table \ref{tab: ablation studies}, we have the following observations:
\textbf{1)} By comparing MoleculeSTM and Ablation 1, we observe that augmenting molecule-text consistently brings performance gain, indicating that ``scale-is-everything'' perspective might also be applicable and beneficial within the MoLM community.
\textbf{2)} Additionally, a comparison between Ablation 1 and Ablation 2 reveals that incorporating a $S^{2}P$ loss can further enhance the performance. This indicates that aligning molecules and text based on the structural similarity among molecules can effectively mitigate issues related to false positives.
\textbf{3)} Lastly, it is noted that the $ER$ loss uniformly enhances performance across all datasets, with a particularly notable improvement in the ATC dataset. 
This improvement can be attributed to the nature of the ATC dataset, which comprises labels for a molecule classification system and is inherently abstract. 
Consequently, reconstructing missing expertise from such abstract descriptions proves advantageous for model performance.


\smallskip
\noindent \textbf{Sensitivity analysis on $k$.}
In Figure \ref{fig: sensitivity analysis k}, we illustrate how model performance fluctuates depending on $k$, the hyperparameter that specifies the number of molecules sharing the same description. 
We note that as $k$ is reduced ($k = 10$), the count of molecule-text pairs diminishes, leading to a slight decline in model performance. 
Conversely, as $k$ increases ($k = 100$), the issue of false positives intensifies, suggesting that an optimal level of $k$ needs to be determined during training. 
Despite these variations, we observe that \proposed~consistently surpasses MoleculeSTM across all values of $k$, showcasing the robustness of \proposed~against changes in the hyperparameter $k$.

In summary, our model effectively augments molecule-text pair by incorporating chemically viable molecules and employing a $S^{2}P$ loss.
Additionally, transferring the knowledge through $ER$ loss consistently enhances \proposed~performance, particularly with brief and abstract descriptions.
Furthermore, we note the model's robust performance across different levels of $k$, reducing the need for extensive trial-and-error in hyperparameter tuning.

\subsection{Zero-Shot Question and Answering}
\label{sec: Zero-Shot Question and Answering}

\begin{table}
\centering
\caption{Model accuracy (\%) in zero-shot question and answering task.}
    \resizebox{0.99\linewidth}{!}{
    \begin{tabular}{lccccccccc}
    \toprule
    & \multicolumn{4}{c}{\textbf{SMILES}} & & \multicolumn{3}{c}{\textbf{Graph}} \\
    \cmidrule{2-5} \cmidrule{7-9}
    & \multirow{2}{*}{MolT5} & \multirow{2}{*}{BioT5} & \multirow{2}{*}{KV-PLM} & Molecule &  & \multirow{2}{*}{MoMu} & Molecule & \multirow{2}{*}{\proposed}\\
    &  &  &  & STM & &  & STM & \\
    \midrule
    Descr. & 24.84 & 27.54 & 30.07 & 36.11 &  & 38.31 & 37.97 & \textbf{39.26}  \\
    Pharma. & 22.49 & 27.03 & 26.68 & 29.60 &  & 29.85 & 29.52 & \textbf{31.58}  \\
    \bottomrule
    \end{tabular}}
    \label{tab: question answering}
\end{table}

\begin{table}
\centering
\caption{Model accuracy (\%) after adopting \proposed~training strategy to MolCA.}
    \resizebox{0.99\linewidth}{!}{
    \begin{tabular}{lcccccccccc}
    \toprule
    & \multicolumn{3}{c}{\textbf{Given Molecule @ 20}} & & \multicolumn{3}{c}{\textbf{Given Text @ 20}} & & \multicolumn{2}{c}{\textbf{Q\&A}} \\
    \cmidrule{2-4} \cmidrule{6-8} \cmidrule{10-11}
    & Descr. & Pharma. & ATC & & Descr. & Pharma. & ATC & & Descr. & Pharma.\\
    \midrule
    MolCA & 97.93 & 83.71 & 59.01 &  & 98.06 & 83.50 & 55.86 & & 41.93 & 33.09 \\
    + \proposed & \textbf{98.46} & \textbf{85.45} & \textbf{59.34} &  & \textbf{98.51} & \textbf{84.93} & \textbf{57.53} &  & \textbf{43.25} & \textbf{34.43} \\
    \bottomrule
    \end{tabular}}
    \label{tab: MolCA}
\end{table}

\begin{table*}[t]
\centering
\caption{ROC-AUC performance in molecular property prediction task. The value within the brackets indicates the variance observed across five trials. Bold text denotes the top performance, while an underline highlights the second best. ''Avg. Rank" represents the average ranking across all datasets.}
    \resizebox{0.85\linewidth}{!}{
    \begin{tabular}{lccccccccc}
    \toprule
     & \textbf{BBBP} & \textbf{Tox21} & \textbf{ToxCast} & \textbf{Sider} & \textbf{Clintox} & \textbf{MUV} & \textbf{HIV} & \textbf{Bace} & \textbf{Avg. Rank} \\
    \midrule
    \multicolumn{10}{l}{\textbf{GraphSSL}} \\
    \midrule
    AttrMask & 68.92 \footnotesize{(1.68)} & 74.86 \footnotesize{(0.64)}& 64.12 \footnotesize{(0.22)} & 59.56 \footnotesize{(1.51)} & 86.52 \footnotesize{(0.53)} & 75.68 \footnotesize{(1.71)} & 75.79 \footnotesize{(1.06)} & 78.71 \footnotesize{(2.47)} & 5.50 \\
    ContextPred & 66.77 \footnotesize{(1.39)} & 73.95 \footnotesize{(0.42)}& 61.77 \footnotesize{(0.67)} & 54.51 \footnotesize{(2.11)} & 81.45 \footnotesize{(3.78)} & 72.88 \footnotesize{(1.80)} & 66.51 \footnotesize{(3.74)} & 74.94 \footnotesize{(6.34)} & 7.63 \\
    GPT-GNN & 60.74 \footnotesize{(0.32)} & 72.14 \footnotesize{(0.55)}& 59.55 \footnotesize{(0.51)} & 54.69 \footnotesize{(0.31)} & 55.87 \footnotesize{(1.92)} & 71.74 \footnotesize{(1.01)} & 71.20 \footnotesize{(0.52)} & 73.23 \footnotesize{(2.63)} & 8.63 \\
    InfoGraph & 66.28 \footnotesize{(2.12)} & 73.12 \footnotesize{(0.49)}& 61.52 \footnotesize{(0.56)} & 57.82 \footnotesize{(1.86)} & 89.14 \footnotesize{(3.30)} & 73.94 \footnotesize{(3.72)} & 77.14 \footnotesize{(1.32)} & 69.41 \footnotesize{(0.39)} & 6.38 \\
    GraphMVP & \underline{69.86} \footnotesize{(0.90)} & 75.37 \footnotesize{(0.20)}& \textbf{65.21} \footnotesize{(0.26)} & 60.12 \footnotesize{(0.60)} & 87.98 \footnotesize{(1.46)} & 76.61 \footnotesize{(0.91)} & 76.12 \footnotesize{(1.04)} & 79.30 \footnotesize{(1.17)} & 3.25 \\
    Mole-BERT & 65.50 \footnotesize{(1.19)} & 74.05 \footnotesize{(0.52)}& 64.75 \footnotesize{(0.71)} & 57.09 \footnotesize{(1.05)} & \textbf{92.03} \footnotesize{(1.06)} & 73.95 \footnotesize{(1.41)} & 76.26 \footnotesize{(0.67)} & 76.93 \footnotesize{(0.76)} & 5.50 \\
    \midrule
    \multicolumn{10}{l}{\textbf{MoLM w/ Graph}} \\
    \midrule
    MoMu & 69.70 \footnotesize{(0.47)} & 75.16 \footnotesize{(0.34)}& 64.90 \footnotesize{(0.26)} & \underline{60.21} \footnotesize{(0.76)} & 86.20 \footnotesize{(0.97)} & \underline{76.63} \footnotesize{(1.02)} & 77.10 \footnotesize{(1.00)} & 78.78 \footnotesize{(0.90)} & 3.75 \\
    MoleculeSTM & 69.08 \footnotesize{(0.54)} & \underline{75.47} \footnotesize{(0.29)}& 64.94 \footnotesize{(0.51)} & 59.60 \footnotesize{(0.51)} & 88.46 \footnotesize{(0.99)} & 75.77 \footnotesize{(1.19)} & \underline{77.96} \footnotesize{(0.63)} & \underline{80.10} \footnotesize{(1.16)} & \underline{3.13} \\
    \midrule
    \proposed & \textbf{69.94} \footnotesize{(0.84)} & \textbf{76.19} \footnotesize{(0.27)}& \underline{65.03} \footnotesize{(0.27)} & \textbf{60.69} \footnotesize{(0.70)} & \underline{89.94} \footnotesize{(0.96)} & \textbf{76.76} \footnotesize{(0.96)} & \textbf{78.42} \footnotesize{(0.71)} & \textbf{80.26} \footnotesize{(1.80)} & \textbf{1.25} \\
    \bottomrule
    \end{tabular}}
    \label{tab: molecular property prediction}
\end{table*}

\textbf{Task Description.}
While earlier studies have assessed the retrieval capabilities of MoLM by contrasting the correct description with randomly chosen descriptions from other molecules \cite{liu2023multi,su2022molecular}, we propose a novel task, termed the ``Zero-shot Question and Answering'' task.
Specifically, given a textual description of a molecule, we instruct GPT-4 \cite{achiam2023gpt} to generate a multiple-choice question comprising five options, all derived from the given textual description.
Then, with a generated question and its five corresponding options, we merge the question with each option to form a single input, i.e., $\texttt{input}_{i} = \text{Concat}(\texttt{question},~\texttt{option}_{i})$, where $i = 1, \ldots, 5$.
Given a molecule and these combined inputs, we then select the one that includes the correct answer from the options.
We generate questions based on the textual descriptions used for the cross-modal retrieval task in Section \ref{sec: Zero-Shot Cross-Modal Retrieval}.
Since the ATC dataset consists of brief labels for molecules, we restrict our question generation to descriptions and pharmacodynamics datasets.
Note that the only difference between the options comes from $\texttt{option}_{i}$, which makes the task much harder compared to the cross-modal retrieval task in Section \ref{sec: Zero-Shot Cross-Modal Retrieval}.
In Appendix \ref{app: dataset zero-shot question and answering}, we offer details for generating processes and statistics for questions and answer datasets.

\smallskip
\noindent \textbf{Empirical Results.}
In Table \ref{tab: question answering}, we have the following observations:
\textbf{1)} Comparing to Table \ref{tab: cross retrieval}, we observe that despite having far fewer options for retrieval, most models learning representation space (i.e., KV-PLM, MoMu, MoleculeSTM, and \proposed) perform much worse in the task.
This is because the model must discern based solely on the subtle differences between the \texttt{option}s provided, requiring the model to have a more fine-grained understanding of molecules than cross-modal retrieval.
\textbf{2)} On the other hand, \proposed~consistently outperforms baseline models in this task, showcasing its ability to extract more intricate information from the slight variations in options through the inference of related expertise.
In conclusion, we posit that \proposed~offers benefits in tasks that demand a more intricate understanding of molecules, achieved by integrating additional expertise.

\smallskip
\noindent \textbf{Adaptability to Various Model Architectures.}
In this section, we examine whether the training strategy of \proposed~can be adapted to various model architectures beyond CLIP style architecture \cite{radford2021learning}, i.e., MolCA \cite{liu2023molca}.
MolCA proposes to train Q-Former with three loss terms, i.e., molecule-text contrasting, molecule-text matching, and molecule captioning.
To implement our training strategy, we first increase the number of molecule-text pairs with the proposed sharing strategy.
Then, we replace the molecule-text contrasting loss with a structural similarity preserving ($S^{2}P$) loss, and also incorporate expertise reconstruction ($ER$) loss along with the proposed three losses.
In Table \ref{tab: MolCA}, we observe consistent performance enhancements of the Q-Former architecture when trained with the proposed strategy, highlighting \proposed's broad applicability beyond CLIP-style architecture.

\subsection{Molecular Property Prediction}
\label{sec: Molecular Property Prediction}

\textbf{Task Description.}
In this task, we assess the potential benefits of incorporating external knowledge, i.e., textual descriptions, into the molecule encoder $f_{\text{mol}}$ as done in~\proposed, in enhancing the prediction of molecular properties.
We mainly compare to recent graph self-supervised learning (\textbf{GraphSSL}) approaches, i.e., AttrMask \cite{hu2019strategies}, ContextPred \cite{hu2019strategies}, GPT-GNN \cite{hu2020gpt}, InfoGraph \cite{sun2019infograph}, GraphMVP \cite{liu2021pre}, Mole-BERT \cite{xia2022mole}, and MoLMs which represent molecules with graph structure (\textbf{MoLM w/ Graph}).
Following previous works \cite{liu2021pre,liu2023multi}, we pre-train the molecule encoder $f_{\text{mol}}$ using each of the proposed strategies with the same pre-taining data and fine-tuning on MoleculeNet benchmark \cite{wu2018moleculenet}.
During the fine-tuning stage, we initially divide the fine-tuning dataset according to scaffold information using an 8:1:1 ratio for the training, validation, and test sets, respectively.
Subsequently, we fine-tune the molecular encoder $f_{\text{mol}}$ using the training data across 100 epochs. 
Following previous work \cite{liu2023multi}, the model's performance is then evaluated on the test set where the hyperparameters achieve optimal performance on the validation set.
We run five individual experiments and report the average and standard deviation of the results in Table \ref{tab: molecular property prediction}.
We provide details on the MoleculeNet dataset in Appendix \ref{app: dataset molecular property prediction}.

\smallskip
\noindent \textbf{Empirical Results.}
In Table \ref{tab: molecular property prediction}, we have following observations:
\textbf{1)} We observe that incorporating external textual descriptions into the pre-training phase generally enhances the prediction of molecular properties, as evidenced by improved overall performance (i.e., averaged rank across all eight tasks). 
This improvement is credited to the implicit influence of external domain knowledge, i.e., the textual descriptions of molecules, which is proved to be beneficial for property prediction. \cite{liu2023multi}.
\textbf{2)} Among the MoLM, \proposed~outperforms baseline methods on six out of eight tasks, demonstrating that the knowledge can be more efficiently transferred to molecular property prediction through our strategy.
In summary, our methodology is advantageous not just for cross-modal tasks, but also for tasks within a single modality, demonstrating the versatility of \proposed~across a wide range of downstream applications.

\subsection{Zero-Shot Virtual Screening}
\label{sec: Zero-Shot Virtual Screening}

\begin{table*}
\begin{minipage}{0.7\linewidth}{
\centering
\caption{Utilized \textcolor{Blue}{abstractive} and \textcolor{BrickRed}{detailed} textual descriptions for each dataset.}
    \resizebox{0.98\linewidth}{!}{
    \begin{tabular}{c|c}
    \toprule
    \textbf{Dataset} & \textbf{Textual Description}\\
    \midrule
    \multirow{3}{*}{\textbf{HIA}} & \textcolor{Blue}{Human intestinal absorption (HIA)}\\
    \cmidrule{2-2}
    & \textcolor{BrickRed}{The molecule is positive w.r.t. a property that is defined as 'the ability of the body to be absorbed} \\
    & \textcolor{BrickRed}{ from the human gastrointestinal system into the bloodstream of the human body'} \\
    \midrule
    \multirow{3}{*}{\textbf{Pgp Inhibition}} & \textcolor{Blue}{P-glycoprotein Inhibition} \\
    \cmidrule{2-2}
    & \textcolor{BrickRed}{This molecule is known to inhibit P-glycoprotein, which is an ABC transporter protein involved in intestinal absorption,} \\
    & \textcolor{BrickRed}{drug metabolism, and brain penetration, and its inhibition can seriously alter a drug's bioavailability and safety.} \\
    \midrule
    \multirow{3}{*}{\textbf{DILI}} & \textcolor{Blue}{Inducing liver injury} \\
    \cmidrule{2-2}
    & \textcolor{BrickRed}{This molecule induces liver injury that is most commonly caused by Amoxicillin }\\
    & \textcolor{BrickRed}{clavulanate isoniazid, and nonsteroidal anti-inflammatory drugs.} \\
    \midrule
    \multirow{3}{*}{\textbf{VDR}} & \textcolor{Blue}{Vitamin D receptor} \\
    \cmidrule{2-2}
    & \textcolor{BrickRed}{This molecule is active w.r.t. Vitamin D receptor. The best pharmacophore hypothesis} \\
    & \textcolor{BrickRed}{contains one hydrogen bond acceptor (A), one hydrogen bond donor (D) and two hydrophobic regions (H).} \\
    \bottomrule
    \end{tabular}}
    \label{tab: virtual screening}
}\end{minipage}
\begin{minipage}{0.29\linewidth}{
    \centering
    \includegraphics[width=0.98\textwidth]{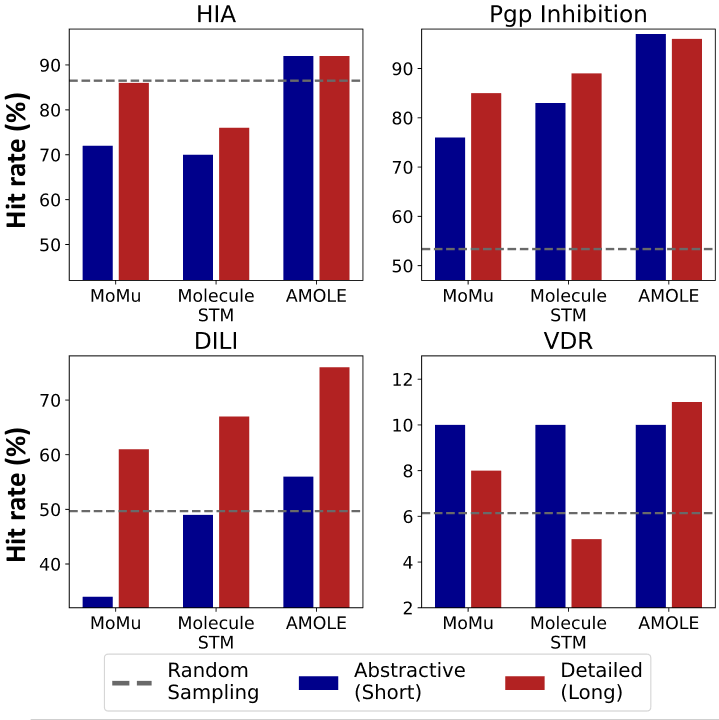} 
    \vspace{-2ex}    
    \captionof{figure}{Hit rate (\%) in zero-shot virtual screening task.}
    \label{fig: virtual screening}
}\end{minipage}
\vspace{-2ex}
\end{table*}

\textbf{Task Description.}
Virtual screening is a computational technique to search large libraries of compounds quickly to identify those structures most likely to have desired properties, emerging as a principal technique in the drug discovery process \cite{mehta2021memes}.
To examine MoLM's real-world applicability, we propose a novel task named ``Zero-shot Virtual Screening,'' in which we assess the model's proficiency in virtual screening drugs by supplying a textual description of a desired property. 
Specifically, given a dataset with binary labels, we first obtain the representation of textual description of the desired property from language model $f_{\text{text}}$.
Then, we identify the top 100 molecules that are closest to the textual representation within the representation space of the dataset and calculate the hit rate to assess the model performance.
Moreover, to further evaluate models' robustness on the various textual descriptions, we conduct the experiment by providing two distinct descriptions for each property as shown in Table \ref{tab: virtual screening}: one brief and abstract, and the other longer and more detailed, offering comprehensive information about the property.
We provide details on each dataset in Appendix \ref{app: dataset VS task}.

\smallskip
\noindent \textbf{Empirical Results.}
In Figure \ref{fig: virtual screening}, we have the following observations:
\textbf{1)} While previous studies have shown comparable results on various downstream tasks, their effectiveness in virtual screening leaves room for improvement. 
Notably, these models often yield results inferior to the mere random selection of molecules (i.e., gray dashed line in Figure \ref{fig: virtual screening}), and their performance significantly fluctuates based on the specific textual description used.
\textbf{2)} However, \proposed~consistently performs the best in various datasets and different types of descriptions, further demonstrating its ability to learn a more qualified joint space between molecules and text.
\textbf{3)} One interesting observation is that, \proposed~demonstrates notably robust performance compared to the baseline methods, regardless of whether the description is abstract or detailed. 
This can be attributed to the expertise transfer module, which equips the model with the ability to deduce related information even when only an abstract level of detail is provided.

\begin{figure}[t]
    \centering
    \includegraphics[width=0.8\columnwidth]{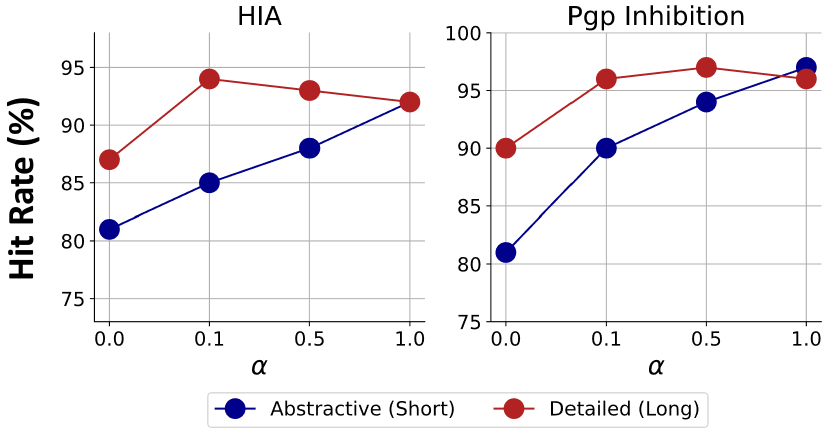} 
    \vspace{-1ex}
    \caption{Sensitivity analysis on $\alpha$.}
    \label{fig: sensitivity analysis}
    \vspace{-5ex}
\end{figure}

\smallskip
\noindent \textbf{Sensitivity Analysis on $\mathbf{\alpha}$.}
In this section, we further investigate how the expertise reconstruction ($ER$) loss weighting factor, i.e., $\alpha$, affects the model robustness on virtual screening performance.
In Figure \ref{fig: sensitivity analysis}, we observe that the model's performance remains largely consistent when provided with a detailed and lengthy prompt, its performance on abstract and brief prompts significantly varies with the choice of $\alpha$. 
Specifically, the model's effectiveness on abstract and concise prompts improves as $\alpha$ increases. 
This suggests that the expertise transfer module effectively allows the model to infer related information, thereby enhancing stability and performance, even when faced with brief and abstract prompts.
In conclusion, \proposed~can consistently screen the qualified molecules given any textual description, highlighting the adaptability of \proposed~for real-world drug discovery process.

\section{Conclusion and Future Works}
In this paper, we introduce \proposed, which tackles two distinct challenges in MoLM that set it apart from VLM: the limited availability of molecule-text paired data in terms of both quantity and expertise.
To address these issues, we propose to selectively share the descriptions among the structurally similar molecules based on their structural similarity, 
and transfer the expertise between molecules by training the model to reconstruct one description from another.
Our extensive evaluation of four downstream tasks, including two innovative and practical tasks such as zero-shot question and answering and zero-shot virtual screening, demonstrate the efficacy of \proposed~in grasping the nuances of molecules and their textual descriptions, highlighting its promising applicability for practical use in the drug discovery domain.
On the other hand, while this paper suggests augmenting the molecule-text pairs by sharing descriptions among similar molecules, this method does not increase the total number of descriptions. 
As an interesting future direction, we suggest potential strategy to augment the absolute number of descriptions involves generating new descriptions based on the recent achievements of decoder-only models like GPT, and adapting the style of one expert to another. 

\smallskip
\noindent \textbf{Acknowledgement.}
This work was supported by the National Research Foundation of Korea(NRF) grant funded by the Korea government(MSIT) (RS-2024-00335098), Ministry of Science and ICT (NRF-2022M3J6A1063021), and by the Institute of Information \& communications Technology Planning \& Evaluation (IITP) grant funded by the Korea government(MSIT) (No.2022-0-00077).

\clearpage

\bibliographystyle{ACM-Reference-Format}
\bibliography{sample-base}


\appendix

\section{Implementation Details}
\label{app: Implementation Details}
In this section, we provide implementation details of \proposed.

\smallskip
\noindent\textbf{Text Encoder}~$f_{\text{text}}$.
Following previous work \cite{liu2023multi}, we use BERT architecture \cite{devlin2018bert} as the text encoder $f_{\text{text}}$, and adapt the SciBERT \cite{beltagy2019scibert} checkpoint to initialize the model parameters \footnote{\url{https://huggingface.co/allenai/scibert_scivocab_uncased}}.
Text encoder contains a total 109,918,464 number of parameters.

\smallskip
\noindent\textbf{Molecule Encoder}~$f_{\text{mol}}$.
We use GIN \cite{xu2018powerful} architecture as a molecular encoder $f_{\text{mol}}$, which has been widely used as the backbone model in recent graph self-supervised learning works \cite{hu2019strategies,liu2021pre}.
Additionally, we initialize the encoder parameters using the GraphMVP \cite{liu2021pre} checkpoints provided by the original authors \footnote{\url{https://huggingface.co/chao1224/MoleculeSTM/tree/main/pretrained_GraphMVP}}.
Molecule encoder contains a total 1,885,206 number of parameters.

\smallskip
\noindent\textbf{Training Details.}
Our method is implemented on Python 3.7.16, PyTorch 1.10.1, and Torch-geometric 2.0.3. All experiments are conducted using an 80GB NVIDIA A100 GPU.
It takes 90 minutes per epoch for training, a total of 2700 minutes for training.

\smallskip
\noindent\textbf{Hyperparameters.}
We list the key hyperparameters used during training in Table \ref{app tab: hyperparameters}.

\begin{table}[h]
\centering
\small
    \begin{tabular}{lc}
    \toprule
    Hyperparameter & Value \\
    \midrule
    Training epochs & 30\\
    Learning rate for text encoder $f_\text{text}$ & 1e-5 \\
    Learning rate for molecule encoder $f_\text{mol}$ & 1e-5 \\
    Temperature for pseudo label $\tau_{1}$ & 0.1 \\
    Temperature for model prediction $\tau_{2}$ & 0.1 \\
    Number of similar molecules $k$ & 50 \\
    Replacement ratio for original molecule $p$ & \{0.2, 0.5\} \\
    Weight of expertise reconstruction loss $\alpha$ & \{0.1, 1.0\} \\
    \bottomrule
    \end{tabular}
    \caption{Hyperparameter specifications for \proposed~pretraining.}
    \label{app tab: hyperparameters}
\end{table}

\section{Baseline Methods}
\label{app: Baseline Methods}
In this section, we elaborate on baseline methods compared during the experiments. 

\smallskip
\noindent \textbf{Single Encoder Models.}
For single encoder models, we mainly compare \proposed~with MolT5, BioT5, and KV-PLM.
While T5-based models were not originally created for capturing the representations of molecules and text descriptions, we evaluate their performance by leveraging the hidden representations, aligning with approaches used in prior studies \cite{seidl2023enhancing}.
\begin{itemize}[leftmargin=.1in]
    \item \textbf{MolT5} \cite{edwards2022translation} stands out as a trailblazer in the field of molecule language models. It introduces a novel approach of pre-training the model on an extensive dataset of unlabeled natural language texts and molecular strings (SMILES), using a denoising objective. Subsequently, the model undergoes fine-tuning for tasks such as molecule captioning and generation.
    For evaluation, we utilize the checkpoints provided by the authors, which are available in the Huggingface repository \footnote{\url{https://huggingface.co/laituan245/molt5-large-caption2smiles}}.
    \item \textbf{BioT5} \cite{pei2023biot5} extends MolT5 by incorporating a diverse array of pre-training tasks, such as denoising molecule SELFIES, protein FASTA sequences, general text, wrapped text, as well as translating between bio-sequences and structured text descriptions.
    We also utilize checkpoints provided by the authors, which are available in the Huggingface repository \footnote{\url{https://huggingface.co/QizhiPei/biot5-base-text2mol}}.
    \item \textbf{KV-PLM} \cite{zeng2022deep} introduces a pre-training approach for language models that incorporates masked language modeling on a specialized corpus featuring inserted SMILES strings.
    We utilize the checkpoint available on the author's Github repository \footnote{\url{https://github.com/thunlp/KV-PLM?tab=readme-ov-file}}.
\end{itemize}

\smallskip
\noindent \textbf{Separate Encoder Models.}
For separate encoder models, we mainly compare \proposed~with MoleculeSTM and MoMu.
To isolate the impact of the training loss on model performance, we ensure that all models are trained under uniform conditions, using the same training data and model architecture. 
This approach guarantees that any differences in performance can be attributed to the distinct training losses employed by each model.
\begin{itemize}[leftmargin=.1in]
    \item \textbf{MoleculeSTM} \cite{liu2023multi} suggests developing representations for molecules and texts through contrastive learning, where a molecule and its corresponding text are considered a positive pair, and all other combinations within the same batch are treated as negative pairs. Additionally, the authors introduce the extensive dataset, named PubChemSTM, which includes 250K distinct molecules and 281K molecule-text pairs.
    They introduce two distinct models, one utilizing SMILES representations of molecules as model input and the other as graph representations. For the SMILES model, they initially use the MegaMolBART \cite{irwin2022chemformer} checkpoint, and for the graph model, the GraphMVP \cite{liu2021pre} checkpoint. 
    However, due to issues with the CUDA environment, we utilize ChemBERTa \cite{chithrananda2020chemberta} \footnote{\url{https://huggingface.co/seyonec/ChemBERTa-zinc-base-v1}} as the initial checkpoint for the SMILES model training and GraphMVP \footnote{\url{https://huggingface.co/chao1224/MoleculeSTM/tree/main/pretrained_GraphMVP}} for the graph model in MoleculeSTM.
    \item \textbf{MoMu} \cite{su2022molecular} introduces a method of contrasting among molecules themselves with an additional loss function besides contrasting molecules and texts.
\end{itemize}

\smallskip
\noindent \textbf{Graph Self-Supervised Learning Methods.}
In Section \ref{sec: Molecular Property Prediction}, we compare \proposed~to representative graph self-supervised learning approaches, which will be briefly introduced here.
\begin{itemize}[leftmargin=.1in]

    \item Attribute Masking (\textbf{AttrMask}) \cite{hu2019strategies} introduces a technique of randomly masking the attributes of nodes or edges in the input and pre-training a GNN to predict these masked attributes.
    \item Context Prediction (\textbf{ContextPred}) \cite{hu2019strategies} proposes to pre-train GNN to ensure nodes situated in analogous structural contexts are represented by proximate embeddings by using subgraphs to predict their surrounding graph structures.
    \item \textbf{GPT-GNN} \cite{hu2020gpt} introduces a method for pre-training GNN through the generation of attributed graphs, which is achieved by dividing the generation process into two distinct phases: the generation of node attributes and the formation of edges between nodes.
    \item \textbf{InfoGraph} \cite{sun2019infograph} proposes to maximize the mutual information between graph-level representation and node-level representation via contrastive learning.
    \item \textbf{GraphMVP} \cite{liu2021pre} introduces a hybrid training approach that merges generative and contrastive methodologies. 
    In the contrastive setting, 2D molecular graphs and their corresponding 3D structures are considered positive pairs, while all other combinations are treated as negative pairs. 
    On the generative side, the model aims to reconstruct the representation of a 3D structure from its 2D molecular graph counterpart and vice versa, facilitating a comprehensive understanding of molecular structures from both dimensions.
\end{itemize}

\section{Datasets}
\label{app: Datasets}
\begin{figure*}[t!]
    \centering
    \includegraphics[width=0.99\textwidth]{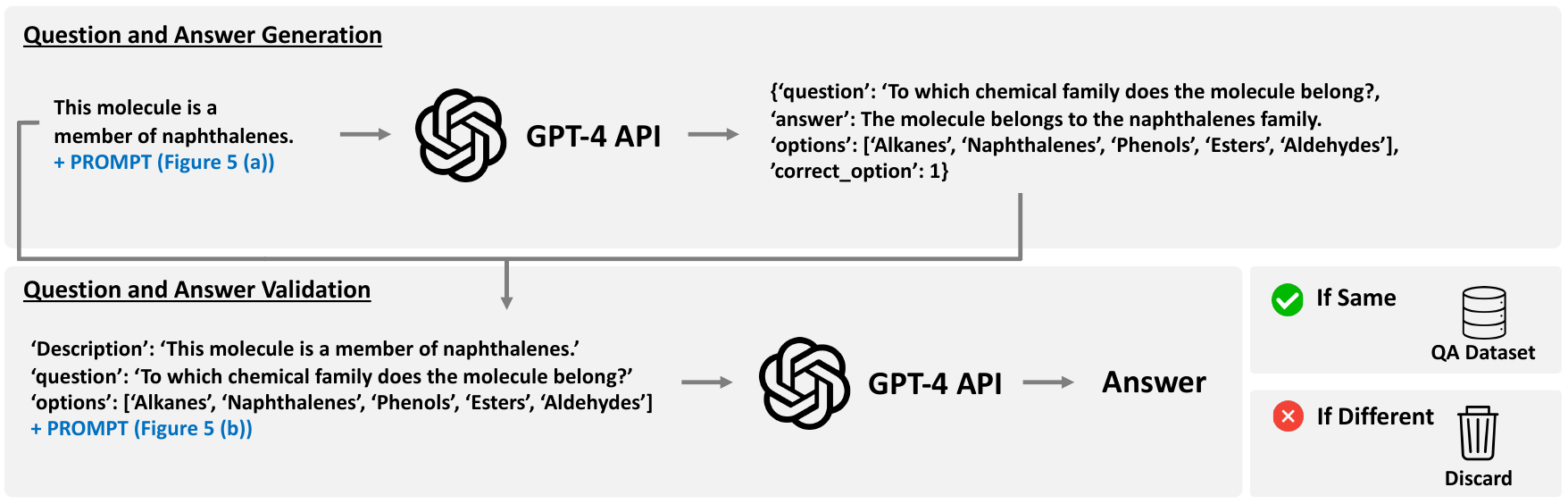} 
    \caption{Overall pipeline for generating and validating questions and answering datasets.}
    \label{app fig: pipeline QAs}
\end{figure*}

\begin{figure*}[t!]
    \centering
    \includegraphics[width=0.99\textwidth]{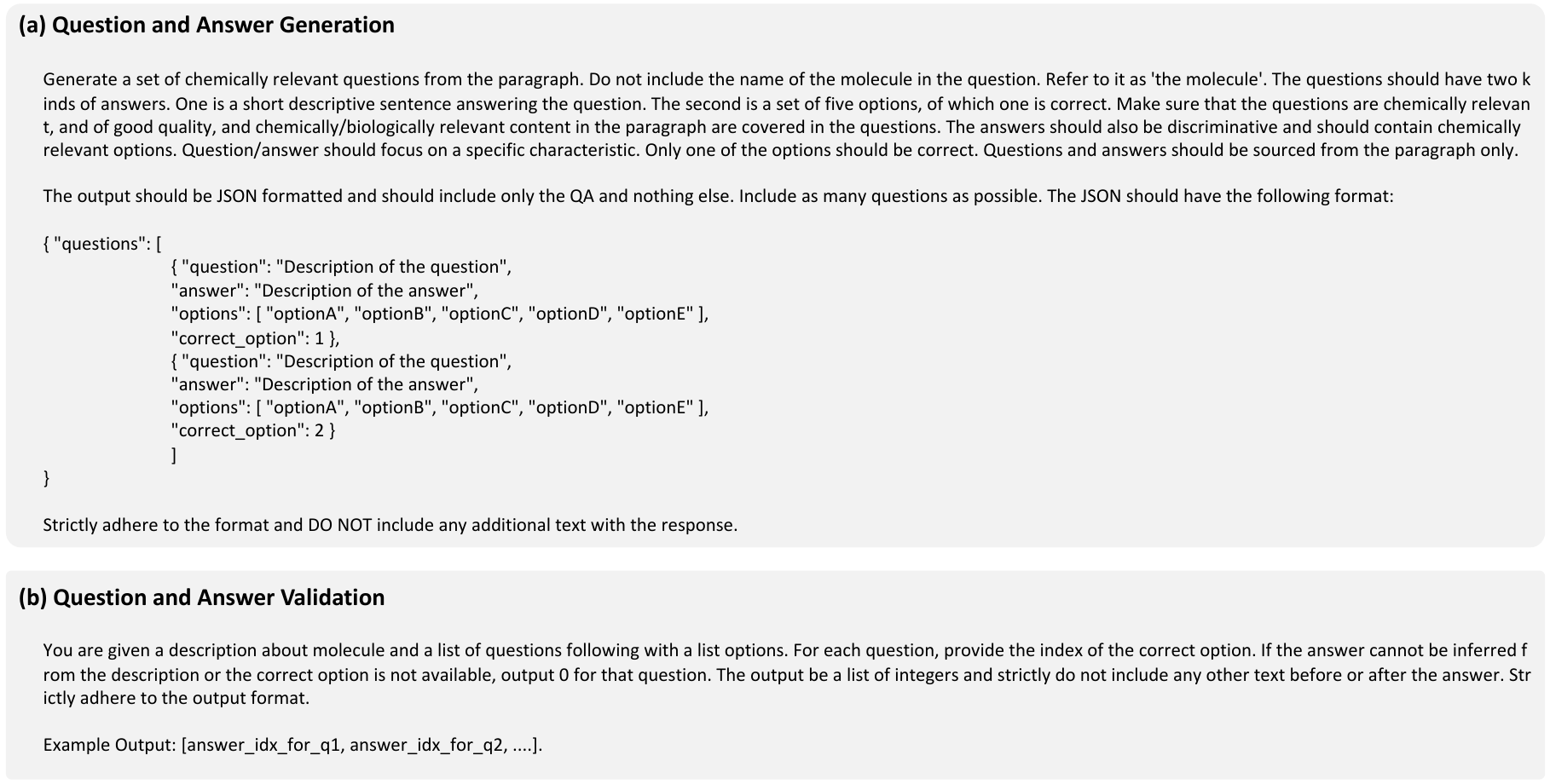} 
    \caption{Prompts for (a) generating and (b) validating question and answer from textual description of molecule.}
    \label{app fig: generating validating QAs}
\end{figure*}

\begin{figure*}[t!]
    \centering
    \includegraphics[width=0.99\linewidth]{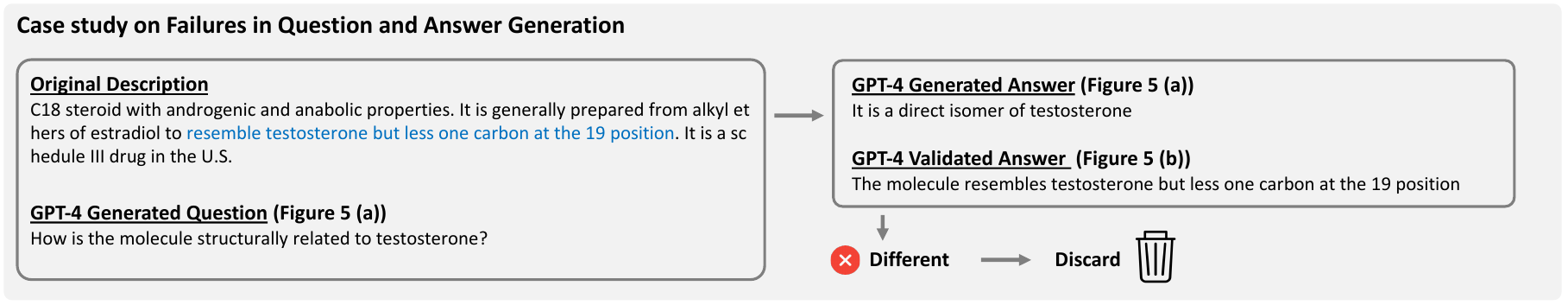} 
    \caption{Failure cases in generating questions and answers.}
    \label{app fig: failure case QAs}
\end{figure*}

\begin{figure*}[t!]
    \centering
    \includegraphics[width=0.95\linewidth]{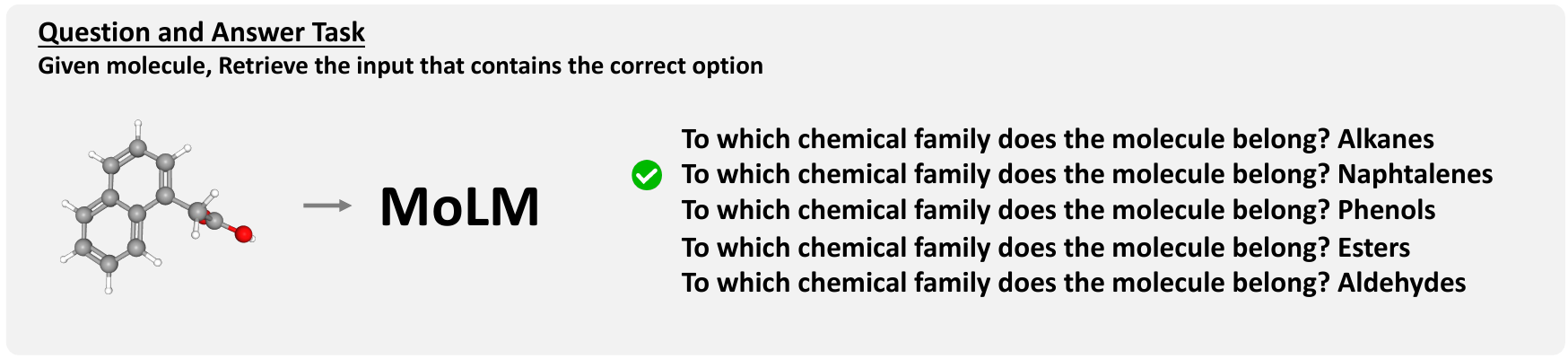} 
    \caption{Evaluation scheme for question and answering task.}
    \label{app fig: evaluation scheme QAs}
\end{figure*}

\subsection{Zero-Shot Cross-Modal Retrieval}
\label{app: dataset zero-shot cross-modal retrieval}
Following previous work \cite{liu2023multi}, we use the datasets extracted from the Description field, the Pharmacodynamics field, and the anatomical therapeutic chemical (ATC) field in the DrugBank database \cite{wishart2018drugbank}.
Each field contains the following information:
\begin{itemize}[leftmargin=.1in]
    \item \textbf{Description (Descr.)} field provides an overview of \textbf{1,154} drugs, including their chemical characteristics, development history, and standing in terms of regulatory approval.
    \item  \textbf{Pharmacodynamics (Pharma.)} field explores the effects and mechanisms of \textbf{1,005} drugs on the body, detailing the biochemical and physiological interactions and responses induced by the drug within the organism.
    \item \textbf{ATC} field represents a classification framework that organizes \textbf{3,007} molecules based on the organ or system they target and their therapeutic, pharmacological, and chemical characteristics.
\end{itemize}
We use the datasets provided in the repository from MoleculeSTM \footnote{\url{https://huggingface.co/datasets/chao1224/MoleculeSTM/tree/main/DrugBank_data/raw}} for our evaluation purposes.
It is worth noting that, we ensure that molecules appearing in the training dataset with identical canonical SMILES are excluded to avoid data leakage.
Moreover, for the ATC dataset, exclusion criteria also consider high similarity between textual descriptions in addition to identical canonical SMILES.

\subsection{Zero-Shot Question and Answering}
\label{app: dataset zero-shot question and answering}
We generate questions based on the textual descriptions used for the cross-modal retrieval task in Section \ref{sec: Zero-Shot Cross-Modal Retrieval}.
Please note that we restrict our question generation to descriptions and pharmacodynamics datasets since the ATC dataset consists of brief labels for molecules.

\smallskip
\noindent\textbf{Question and Answer Generation.}
In Section \ref{sec: Zero-Shot Question and Answering}, we assess the MoLM's capacity to discern the correct answer from options with minor differences, aiming to evaluate a more nuanced understanding of molecules and their textual descriptions. 
To achieve this, we employ GPT-4 to craft a multiple-choice question with five options, each based on the textual descriptions of molecules, by providing specific prompts to guide its generation.
We provide the precise prompts used for generating questions and answers (QAs) in Figure \ref{app fig: generating validating QAs} (a).
Consequently, we acquire a sum of 8,215 QA pairs for the description dataset and 7,300 QA pairs for the pharmacodynamics dataset.

\smallskip
\noindent\textbf{Question and Answer Validation.}
While GPT-4 proficiently generates questions and answers based on the textual descriptions of molecules, there are cases where it produces questions with wrong answers.
Consequently, we refine the generated questions and answers by assessing if GPT-4 accurately identifies the same answers as GPT-4 provided using the original textual descriptions and the generated questions.
As an example, given the context of the original description and the GPT-4 generated questions, there are cases where the answers generated by GPT-4 and the answers validated by GPT-4 are different, as shown in Figure \ref{app fig: failure case QAs}.
Since GPT-4 fail to produce consistent answers based on the original context, we consider these instances as unsuccessful question and answer generation, excluding them from our analysis.
We provide the specific prompts for validating QAs in Figure \ref{app fig: generating validating QAs} (b).
After filtering out invalid QA pairs, we obtain a sum of \textbf{7,986 QA pairs} for the description dataset and \textbf{7,184 QA pairs} for the pharmacodynamics dataset.

\subsection{Molecular Property Prediction}
\label{app: dataset molecular property prediction}
For the molecular property prediction task in Section \ref{sec: Molecular Property Prediction}, we use the MoleculeNet benchmark \cite{wu2018moleculenet} for evaluation, which is commonly used for evaluating the machine learning methods for molecular property prediction \cite{liu2021pre}.
The MoleculeNet benchmark encompasses a diverse datasets, each characterized by distinct properties as follows:
\begin{itemize}[leftmargin=.1in]
    \item The blood-brain barrier penetration \textbf{(BBBP)} dataset comprises binary labels for \textbf{2,039} compounds regarding their barrier permeability, addressing a critical challenge in the development of drugs aimed at the central nervous system.
    \item The toxicology in the 21st Century \textbf{(Tox21)} dataset provides qualitative toxicity measurements for \textbf{7,831} compounds across 12 distinct targets.
    \item The \textbf{ToxCast} dataset offers toxicological data collected from more than 600 experiments on \textbf{8,577} compounds.
    \item The side effect resource \textbf{(Sider)} dataset categorizes the side effects of \textbf{1,427} approved drugs into 27 different organ system classes.
    \item The \textbf{Clintox} dataset comprises two classification tasks for \textbf{1,477} drug compounds, focusing on 1) toxicity during clinical trials and 2) FDA approval status.
    \item The \textbf{MUV} dataset features 17 demanding tasks for \textbf{93,087} compounds, curated from the PubChem BioAssay database.
    \item The \textbf{HIV} dataset, created by the Drug Therapeutics Program (DTP) AIDS Antiviral Screen, assesses the capacity of more than \textbf{41,127} compounds to block the replication of HIV.
    \item The \textbf{BACE} dataset offers qualitative binding outcomes for a collection of inhibitors targeting human $\beta$-secretase 1, encompassing \textbf{1,513} compounds.
\end{itemize}

\subsection{Zero-Shot Virtual Screening}
\label{app: dataset VS task}
For the zero-shot virtual screening task, we utilize datasets from the Therapeutics Data Commons (TDC)~\cite{huang2021therapeutics} and LIT-PCBA \cite{tran2020lit}, containing the drugs that exhibit a range of desirable properties.
Among the various datasets, we utilize the following datasets:
\begin{itemize}[leftmargin=.1in]
    \item The human intestinal absorption \textbf{(HIA)} dataset comprises \textbf{578} drugs and their capacity for absorption from the human gastrointestinal tract into the bloodstream.
    \item The P-glycoprotein inhibition \textbf{(Pgp Inhibition)} dataset includes \textbf{1,212} drugs, detailing their Pgp inhibitory activities, which can significantly affect a drug's bioavailability and safety profile.
    \item The drug-induced liver injury \textbf{(DILI)} dataset includes \textbf{475} drugs, annotated with information regarding their potential to induce liver damage.
    \item The Vitamin D Receptor \textbf{(VDR)} dataset initially contains 263,303 drugs, of which 655 are active. Considering the significant imbalance between active and inactive drugs, we sample a subset of 10,000 drugs from the inactive category for analysis, i.e., a total of \textbf{10,655} drugs.
\end{itemize}

\section{Experimental Setups}
\label{app: Experimental Setups}
\subsection{Zero-Shot Cross-Modal Retrieval}
\label{app: experimental setup retrieval task}
In this section, we provide further details on experimental setups for the zero-shot cross-modal retrieval task.
Following previous work \cite{liu2023multi}, we evaluate the task performance in two distinctive settings 1) given molecule to retrieve the textual description, and 2) given texture description to retrieve the molecule.
In each scenario, we conducted experiments with a range of options, specifically 4, 10, and 20 choices. 
Within these options, one is the matching counterpart, while the others are randomly selected from the dataset.
Following this, the model performance is determined by its capacity to pinpoint the correct counterpart from the options provided, such as correctly matching the description to the given molecule or vice versa.
For the evaluation, we conducted five separate experiments, each featuring a distinct random selection of options, and we present both the mean and standard deviation of these experiments.

\subsection{Zero-Shot Question and Answering}
\label{app: experimental setup QA task}
In this section, we provide further details on the zero-shot questions and answering task.
We provide an overall pipeline for generating a dataset for the question and answering tasks in Figure \ref{app fig: pipeline QAs}.
After the creation of a set of validated questions and answers, we design the question and answering task as a retrieval task. 
Specifically, for a given question and its five options, we concatenate the question with each option to generate a singular input, i.e., $\texttt{input}_{i} = \text{Concat}(\texttt{question},~\texttt{option}_{i})$, for $i = 1, \ldots, 5$. 
The MoLM then determines the correct answer by selecting from these inputs the one that correctly matches the question as shown in Figure \ref{app fig: evaluation scheme QAs}.
The only difference between the options comes from $\texttt{option}_{i}$, makes the task much harder compared to the cross-modal retrieval task in Section \ref{sec: Zero-Shot Cross-Modal Retrieval}.

\subsection{Molecular Property Prediction}
\label{app: experimental setup MPP task}
In this section, we provide details on experimental setups for the molecular property prediction task.
In this task, we evaluate how the pre-training methods affect the prediction of various molecular properties.
To achieve this, we initially divide the dataset according to scaffold information using an 8:1:1 ratio for the training, validation, and test sets, respectively.
That is, the molecules in the training, validation, and test sets possess distinct scaffolds.
Subsequently, we fine-tune the molecular encoder $f_{\text{mol}}$ using the training data across 100 epochs. 
Following previous work \cite{liu2023multi}, the model's performance is then evaluated on the test set where the hyperparameters achieve optimal performance on the validation set.
We ran five individual experiments and report the average and standard deviation of the results.

\subsection{Zero-Shot Virtual Screening}
\label{app: experimental setup VS task}

In this task, we assess the model's capability to conduct virtual screening for drugs with specific properties described in a language model prompt. 
Given a dataset with binary labels, we evaluate the model performance by identifying and counting the number of molecules tagged with a positive label among those that have been screened.
We start by encoding the prompt into its representation, followed by identifying the top 100 molecules whose representations are nearest to that of the prompt in the representation space.
For instance, in the case of the HIA dataset, we initially identify the top 100 molecules that are closest to the prompt ''Human intestinal absorption (HIA)" within the representation space among a total of 578 molecules and count the molecules with positive labels.
Additionally, for every dataset, we employ two types of prompts with distinct characteristics, i.e., one with short and abstract descriptions and one with long and detailed descriptions.


\clearpage


\end{document}